\documentclass[12pt,journal,compsoc]{IEEEtran}
\usepackage{graphicx}
\usepackage{algorithm}
\usepackage{algpseudocode}
\usepackage{subcaption}
\usepackage{color}

\ifCLASSOPTIONcompsoc
\else
\fi

\ifCLASSINFOpdf
\else
\fi

\newcommand{\RNum}[1]{\uppercase\expandafter{\romannumeral #1\relax}}

\begin{document}
%
\title{An Empirical Study of Recent \\Face Alignment Methods }

\author{Heng Yang, Xuhui Jia, Chen Change Loy and Peter Robinson
\thanks{Yang and Robinson are with Computer Laboratory, University of Cambridge, England.}
\thanks{Jia is with the University of Hong Kong, Hong Kong.}
\thanks{C.C. Loy is with Multimedia Laboratory, Chinese University of Hong Kong, Hong Kong.}
\thanks{Project: \textcolor{red} {https://www.cl.cam.ac.uk/~hy306/FaceAlignment.html}}
}

\markboth{}%
{Shell \MakeLowercase{\textit{et al.}}: Bare Demo of IEEEtran.cls for Computer Society Journals}


\IEEEtitleabstractindextext{%
\begin{abstract}
The problem of face alignment has been intensively studied in the past years. A large number of novel methods have been proposed and reported very good performance on benchmark dataset such as 300W. However, the differences in the experimental setting and evaluation metric, missing details in the description of the methods make it hard to reproduce the results reported and evaluate the relative merits. For instance, most recent face alignment methods are built on top of face detection but from different face detectors. In this paper, we carry out a rigorous evaluation of these methods by making the following contributions: 1) we proposes a new evaluation metric for face alignment on a set of images, i.e., area under error distribution curve within a threshold, AUC$_\alpha$, given the fact that the traditional evaluation measure (mean error) is very sensitive to big alignment error. 2) we extend the 300W database with more practical face detections to make fair comparison possible. 3) we carry out face alignment sensitivity analysis w.r.t. face detection, on both synthetic and real data, using both off-the-shelf and re-retrained models. 4) we study  factors  that are particularly important to achieve good performance and provide suggestions for practical applications. Most of the conclusions drawn from our comparative analysis cannot be inferred from the original publications.

\end{abstract}

}

\maketitle

\IEEEdisplaynontitleabstractindextext

%
\IEEEpeerreviewmaketitle

\section{Introduction}
%

\IEEEPARstart{T}{he} study of face alignment, or facial landmarks localisation, has made rapid progresses in recent years. Several methods have reported close-to-human performance on benchmark datasets, e.g. the Automatic Facial Landmark Detection in-the-Wild Challenge (300W), of which the images that are acquired from unconstrained environments. However, while there is much ongoing research in computer vision approaches for face alignment, varying evaluation protocols, lack descriptions of critical details and the use of different experimental setting or datasets makes it hard to shed light on how to make an assessment of their cons and pros, and what are the important factors influential to performance. 

\begin{figure}
\centering
\includegraphics[width=0.15\textwidth,height=0.15\textwidth]{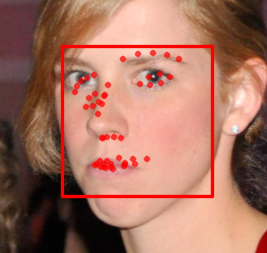}
\includegraphics[width=0.15\textwidth,height=0.15\textwidth]{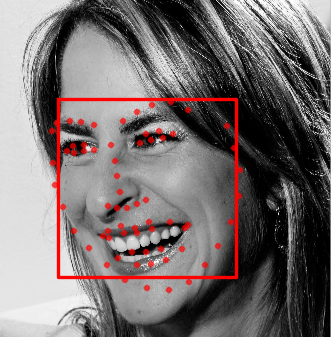}
\includegraphics[width=0.15\textwidth,height=0.15\textwidth]{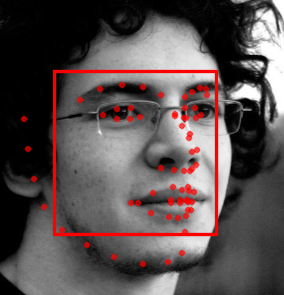}\\
\includegraphics[width=0.15\textwidth,height=0.15\textwidth]{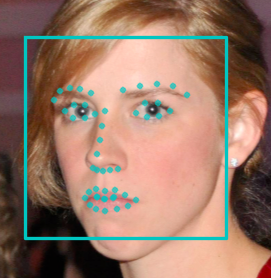}
\includegraphics[width=0.15\textwidth,height=0.15\textwidth]{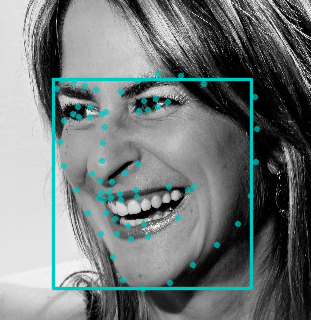}
\includegraphics[width=0.15\textwidth,height=0.15\textwidth]{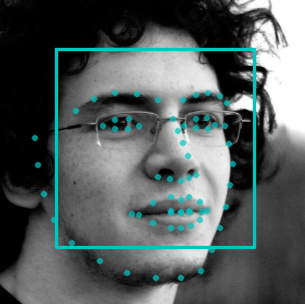}
\caption{Face alignment results given different face detections. The upper row shows the results from HeadHunter \cite{mathias2014face} while the the lower row shows the results from the \textit{best} face detection. From left to right, SDM \cite{xiong2013supervised}, CFAN \cite{cfaneccv2014}, TREES \cite{kazemi2014one}, their corresponding \textit{best} face detection: Viola-Jones \cite{viola2001rapid}, IBUG \cite{sagonas300} and dlib \cite{dlib09}. }
\label{fig::illu}
\end{figure}

\begin{figure}
\includegraphics[ width=0.24\textwidth,height=0.2\textwidth]{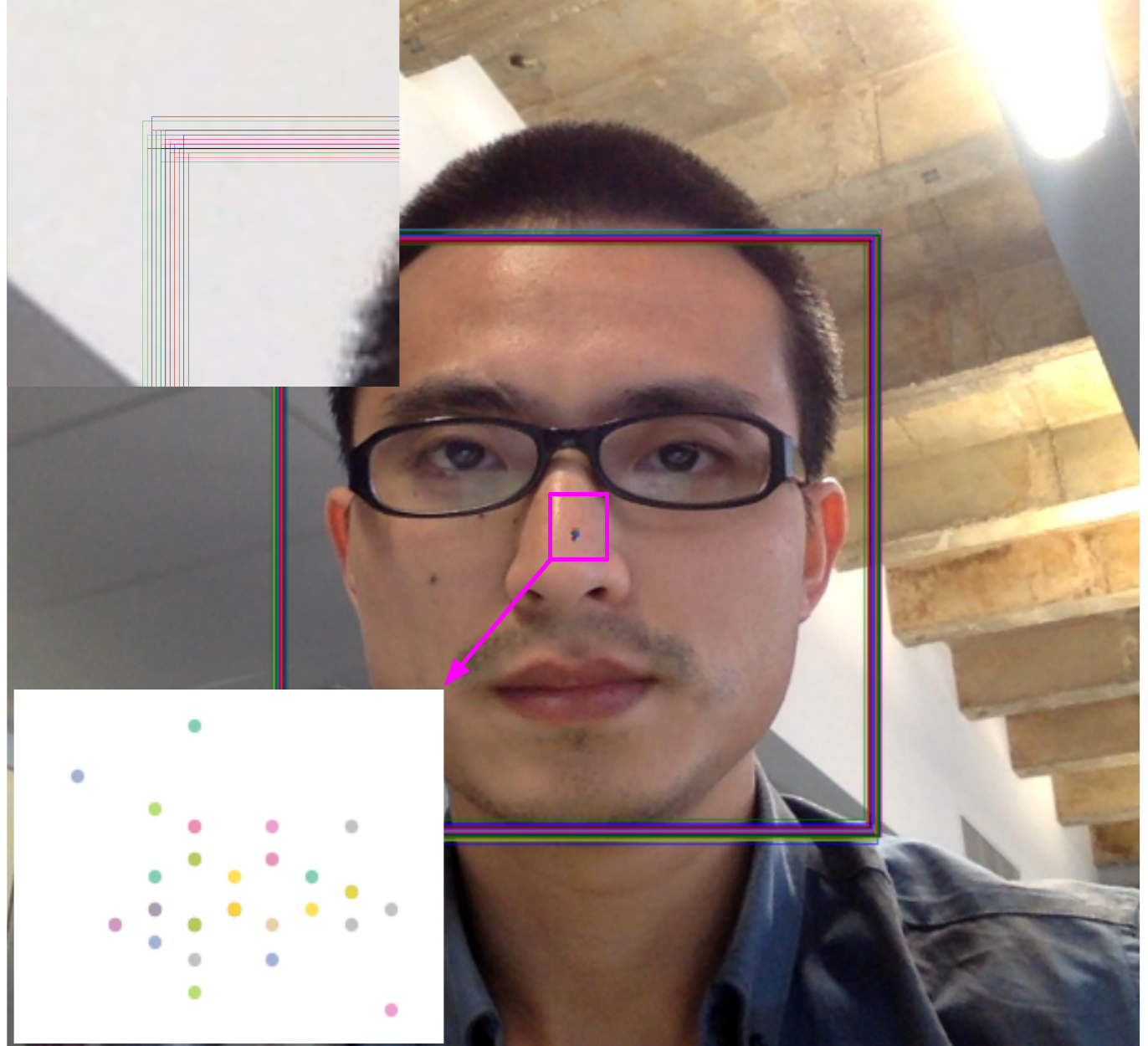}
\includegraphics[ width=0.24\textwidth,height=0.2\textwidth]{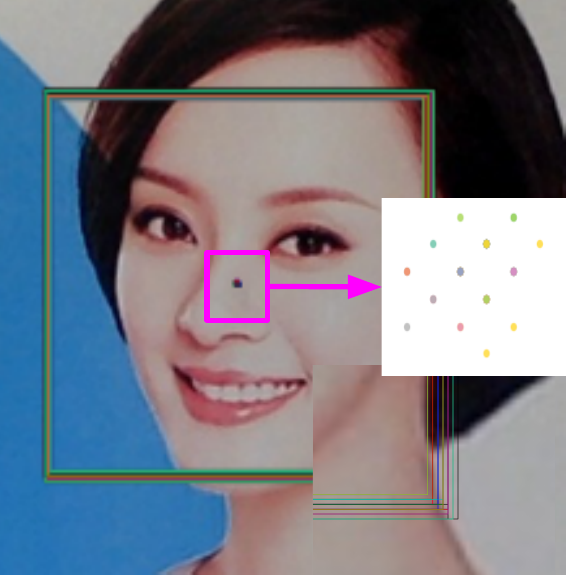}
\caption{Viola-Jones Face detection jitter. Left, sequence of still live face; right, sequence of static face photo.}
\label{fig::facedetjitter}
\end{figure}

Face alignment is often served as an intermediate step in commonly used face analysis pipeline (face detection $\Rightarrow$ face alignment $\Rightarrow$ face recognition). Despite the fact that most current face alignment methods \cite{zhu2015face,zhang2014facial,cfaneccv2014,asthanaincremental,renface,xiong2013supervised} build on top of face detection, few of them have discussed the face alignment sensitivity w.r.t face detection variation. In Fig.~\ref{fig::illu}, we show some face alignment examples of several state of the art models. In the first row, the Headhunter \cite{mathias2014face} is applied for face detection given its state of the art face detection performance. In the second row, face detection is from the \textit{best} face detector. As can be seen, the methods struggle to obtain reasonable alignment results while face detection changes. Moreover, face detection jitter is a very common phenomenon in reality. As shown in Fig.~\ref{fig::facedetjitter}, face detection jitters several pixels in consecutive frames even there is no face movement at all, which might be due to very tiny lighting noise. Thus we believe it is very meaningful to study the face alignment sensitivity w.r.t face detection changes.  

In this paper, we attempt to study recent face alignment methods regarding above mentioned concerns in an empirical manner. We first extend the 300W benchmark dataset and form the 300W++ dataset by providing several types of popular face detections, in addition to its impractical tight face bounding boxes. For the samples with missing face detections, we fulfil them by using a deep ConvNet regression model. We also point out the issue of using overall mean error as an evaluation criterion as it is too sensitive to big erroneous samples and propose a new evaluation criterion, $AUC_{\alpha}$. The extended dataset and the new evaluation criterion will enable future comparison more convenient and consistent. Then we focus on performance evaluation and sensitivity analysis of recent face alignment methods and make the following contributions:

First, we compare the performance of publicly available off-the-shelf face alignment models on the 300W++ dataset, on both their \textit{best} face detection and other face detections. We carry out synthesised experiments by adding artificial noises on the face detection (face centre shifts and scale changes) and study the sensitivity w.r.t face detection (initialisation) variation. In total, we have run more than 1000 groups of experiments on 11 representative methods and demonstrate their relative merits as well as their sensitivity w.r.t initialisation changes.

Second, to have a fair comparison, we re-train several best performing and typical methods by using the same data (training samples and face detection), and by making other settings such as augmentation number as similar as possible. Then we compare their alignment performance and their robustness against initialisation changes. In this way, we can have a unbiased look at these methods and draw useful and comparative conclusions. 

Third, we revisit a typical cascaded face alignment, Explicit Shape Regression (ESR) \cite{suncvpr2012}, which brought a breakthrough record of face alignment in both accuracy and speed. In the spirit of the devil is in the detail, we study how the performance is influenced by some important factors like initialisations and number of cascades. Those findings are also useful to other cascaded face alignment methods. 

From our empirical study and in-depth comparison, it is able to have an overall picture of the performance of recent face alignment methods and to identify the aspects of the different constructions which are important for performance and which are not.  Most of the results/conclusions in this paper cannot be inferred from the original publications alone. For further comparison, we will release the source code and 300W++ dataset  and  describe all the implementation details (including some that were omitted in the original publications and that were obtained from personal communications with the authors). With the release of the code and dataset, we wish to encourage: 1) the use the practical 300W++ benchmark to make a fair evaluation of new contribution; 2) the application of AUC$_{\alpha}$ as a evaluation metric over a testing set; 3) sensitivity analysis of a face alignment model.  





\section{Face Alignment Methods}
Papers on face alignment have flourished in recent years. Based on whether a method uses specific detector (local expert) for an individual landmark or not, we roughly group the methods into two categories: local-based methods and holistic-based methods. The former usually has explicit local expert model while the latter does not. We investigate on recent state of the art methods in both categories and review some hybrid methods as well. More holistic-based methods are studied as they are more dominant in recent years. We attempt to include as many as possible of the recent advances, but it is hard to cover all of them due to implementation difficulty and space limit. Readers interested in other representative methods such as the original Active Appearance Models (AAM) \cite{cootes2001active} and others are referred to \cite{wang2014facial, cceliktutan2013comparative}. 

\begin{table*}[]
\centering
\caption{Local-based methods}
\label{tab::localmethods}
\begin{tabular}{lcccccc}
\hline
Local Expert & SVMs \cite{rapp2011multiple,belhumeur2011localizing,Zhou_2013_ICCV} & Dict.L \cite{asthana2013robust}  & SVRs \cite{martinez2012local} & CF \cite{Kiani_2013_ICCV} & RF \cite{cootesECCV2012,dantone2012real,yangiccv2013}   & CCNF \cite{baltruvsaitis2014continuous} \\ 
Shape Model  & CLM \cite{cristinacce2006feature, baltruvsaitis2014continuous, cootesECCV2012}  & Mix.Tree \cite{devacvpr2012face, fowlkesocclusion, Yu_2013_ICCV} & DPM \cite{tzimiropoulos2014gauss} & GraphMatch. \cite{Zhou_2013_ICCV}  & RANSAC \cite{belhumeur2011localizing} & HPM \cite{wu2014hierarchical} \\ \hline
\end{tabular}
\end{table*}
\subsection{Local-based methods}
Local-based methods usually consist of two parts: local experts and spatial shape models. The former describes how image around each facial landmark looks like while the latter describes how face shape varies. There are three main types of local experts: 1) Classification-based approaches , e.g. SVMs \cite{rapp2011multiple,belhumeur2011localizing,vukadinovic2005fully} based on various image features such as Gabor \cite{vukadinovic2005fully}, SIFT \cite{lowe2004distinctive},  Discriminative Response Map Fitting (\textbf{DRMF}) by dictionary learning (Dict.L) \cite{asthana2013robust} and multichannel correlation filter (CF) responses \cite{Kiani_2013_ICCV}; 2) Regression-based approaches like Support Vector Regressors (SVRs) \cite{martinez2012local}, Continuous Conditional Neural Fields (\textbf{CCNF}) \cite{baltruvsaitis2014continuous}; 3) Voting-based approaches, including regression forests based voting \cite{cootesECCV2012,dantone2012real,yangiccv2013} and exemplar based voting \cite{smithnonparametric,shen2013detecting}. One typical shape model is the Constrained Local Model (CLM) \cite{cristinacce2006feature}, which has been widely adapted with various local expert models. There are some other shape models such as RANSAC in \cite{belhumeur2011localizing}, graph-matching in \cite{Zhou_2013_ICCV}, Gaussian Newton Deformable Part Model (\textbf{GNDPM}) \cite{tzimiropoulos2014gauss}, mixture of trees \cite{devacvpr2012face} and Hierarchical Probabilistic Model (HPM) \cite{wu2014hierarchical}. 
\begin{table*}[!htp]
\footnotesize
\setlength{\tabcolsep}{3pt}
\centering
\caption{Holistic-based methods and their properties.}
\label{tab::holisticmethods}
\begin{tabular}{lccccccccc}
\hline
Methods        & ESR \cite{asthanaincremental} &SDM \cite{xiong2013supervised}       & RCPR \cite{burgos2013robust}         &  IFA \cite{asthanaincremental} & TREES  \cite{kazemi2014one}        & CFAN \cite{cfaneccv2014}        & TCDCN \cite{cnnficalcvpr2013} &LBF \cite{renface} & CFSS\cite{zhu2015face}\\
initialization &random & mean pose & random       & mean pose     & mean pose & supervised & supervised& mean pose & random\\
features       &pixel diff.& SIFT      & pixel diff.    &HOG    & pixel diff.       & auto-encoder & pixel& LBF&SIFT\\
regressor      &random ferns & linear     & random ferns & linear  & random trees & linear &   ConvNet& linear &linear   \\
\hline
\end{tabular}
\end{table*}
Local-based method has the advantage of making alignment assessment through local likelihood. However, due to the nature of model design, local-based methods are computationally expensive especially when the number of facial landmarks is high. Moreover, in such a method, it is usually a tricky task to balance the local responses and global constraints. In this paper we select three representative local-based methods, namely the CCNF \cite{baltruvsaitis2014continuous}, the GNDPM \cite{tzimiropoulos2014gauss} and the DRMF \cite{asthana2013robust} for evaluation given their state of the art performance among local-based methods.

\subsection{Holistic-based methods}
\label{sec::methods}
Holistic-based methods have gained higher popularity than local-based methods in recent years. Most of them work in a cascaded way similar to the classical Active Appearance Model (AAM) \cite{cootes2001active}. In such methods, the face shape is often represented as a vector of landmark locations, i.e., $S=(\mathrm{x}_1,...,\mathrm{x}_k,...,\mathrm{x}_K) \in\mathbf{R}^{2K}$, where $K$ is the number of landmarks. $\mathrm{x}_k \in \mathbf{R}^2$ is the 2D coordinates of the $k$-th landmark. Most  current holistic-based methods work in a coarse-to-fine fashion, i.e., shape estimation starts from an initial shape $S^0$ and progressively refines the shape by a cascade of $T$ regressors, $R^{1...T}$. Each regressor refines the shape by producing an update, $\Delta S$, which is added up to the current shape estimate. It is summarized in Algorithm \ref{alg::algorithm1} \cite{dollar2010cascaded}.
\begin{algorithm}
\caption{Cascaded Pose Regression}
\label{alg::algorithm1}
\begin{algorithmic}[1]
    \Require{Image $I$, initial pose $S^0$} 
    \Ensure{Estimated pose $S^T$}
    \For {$t$=1 to $T$}
    \State $f^t = h^t(I,S^{t-1})$\Comment{Shape-indexed features} 
    \State $\Delta S = R^t(f^t)$\Comment{Apply regressor $R^t$}
    \State $S^t = S^{t-1}+\Delta S$\Comment{update pose}
     \EndFor
\end{algorithmic}
\end{algorithm}

 Despite different strategies are proposed in recent years, most of them share the above described framework. They differ from each other mainly in three aspects: strategy of setting initialisation; 2) shape-indexed features; 3) regressor. Feature extraction and regression are usually interdependent. 
 
 There are mainly three initialisation schemes: random, mean pose, and supervised. The \textit{random} method usually selects one or several face shapes from a set of training samples and then rescale them w.r.t the provided face bounding box via similarity transformation. The mean pose initialization method calculates a mean shape within the face box. The supervised scheme usually calculate a initialisation shape by using an auxiliary  model (e.g. ConvNet) that usually takes the image content in the face bounding box as input. 

A large variety of image features are utilized as the shape indexed features, that include grey scale pixel value comparison (e.g., \cite{dollar2010cascaded,suncvpr2012,burgos2013robust,kazemi2014one}), hand-crafted features (like SIFT \cite{lowe2004distinctive} in \cite{xiong2013supervised,zhu2015face,tzimiropoulos2015project} and HOG \cite{dalal2005histograms} in \cite{yang2014face}) and learned features (using Auto-encoders \cite{cfaneccv2014} or ConvNet \cite{cnnficalcvpr2013}). 

The regressors also vary a lot in different methods that include random ferns \cite{dollar2010cascaded,suncvpr2012, burgos2013robust}, random forests \cite{kazemi2014one}, Support Vector Regressor \cite{asthanaincremental,yanlearn}  and Supervised Descent Method (SDM) and its extensions \cite{xiong2013supervised,asthanaincremental, xiong2015global,zhu2015face}. Recently, deep learning framework has also been applied in the problem of face alignment, which usually also works in a holistic and coarse-to-fine manner for instance \cite{cnnficalcvpr2013}. It has been improved by Tasks Constrained Deep Convolutional Network (TCDCN) \cite{zhang2014facial}, which jointly optimises face alignment and correlated tasks such as head pose estimation in a single ConvNet framework.  We list the recent holistic-based methods that are going to be investigated in this work and their corresponding properties in Table~\ref{tab::holisticmethods}. It is interesting to note that the combination of features and regressors are either non-linear features + linear regression or linear features + non-linear regression, due to the fact that the mapping from raw image to face pose is a non-linear process. 

\subsection{Hybrid methods}
There are several other hybrid face alignment approaches such as occlusion detection based methods \cite{fowlkesocclusion,yang2015robust}, combined local and holistic-based method in \cite{alabort2015unifying,yang2015robust},  weakly supervised method \cite{pedersoli2014using}, unified face detection and alignment method \cite{chen2014joint}, Active Pictorial Model \cite{antonakos2015active}, etc. Due to their different setting and limited space, we leave their comparison as future work.

\section{Data Preparation and Evaluation Metric}
\subsection{An extended dataset: 300W++}
300W \cite{sagonas300}, created for Automatic Facial Landmark Detection
in-the-Wild Challenge, has been widely used as a benchmark dataset for face alignment in recent years. As it only provides the training images for the challenge, we follow the experiment setting of recent methods \cite{zhu2015face,renface} for training/testing partition. More specifically, the training part consists of AFW, the training images of LFPW and the training images of HELEN, with 3148 samples in total. The testing set
consists of the test images of LFPW, the test images of HELEN and the images in the IBUG set, with 689 samples in total. 300W provides the ground truth locations of 68 facial landmarks. 

It provides two types of ground truth face bounding boxes. One is the tight bounding box of the facial landmarks, and the other is the detection outputs from the mixture of trees  model \cite{devacvpr2012face}, which is very close to the tight bounding box. In practice, it is difficult to obtain such tight bounding box since the the mixture of trees \cite{devacvpr2012face} model is very slow and not very effective. Furthermore, many publicly available models are using more practical face detectors, e.g. SDM uses Viola-Jones \cite{viola2001rapid} face detector. Many publications are lack of details when comparing to other models. Thus we provide face bounding boxes from several popular face detectors including the Viola-Jones \cite{viola2001rapid} detector from Matlab, the HeadHunter \cite{mathias2014face} and HOG+SVM detector from dlib \cite{dlib09}. The ground truth detection is called IBUG.  

\begin{figure}
\centering
\includegraphics[width=0.5\textwidth,height=0.15\textwidth]{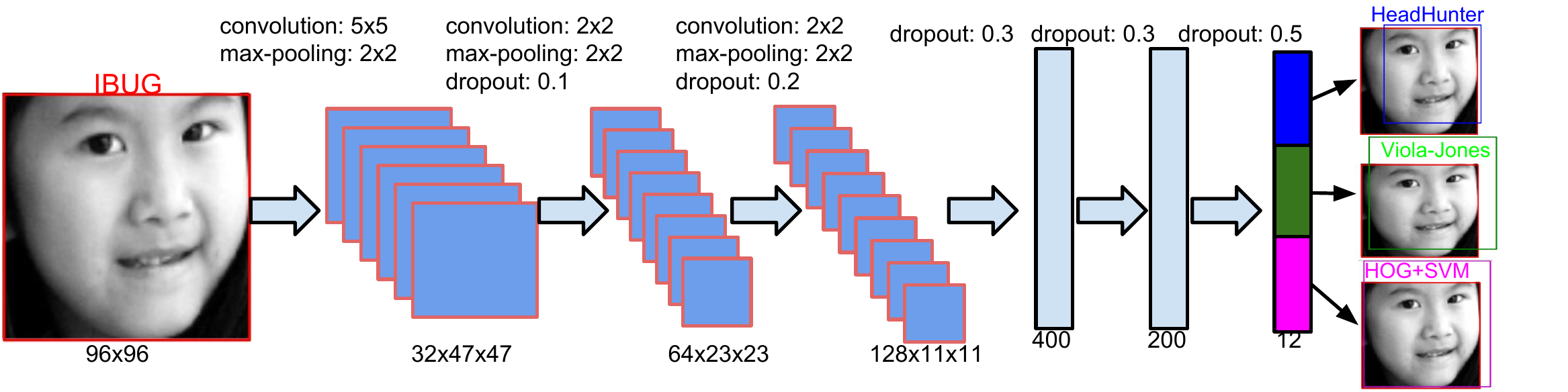}\\
\caption{ConvNet structure specification for missing face detection fulfilment.  }
\label{fig::facebbnet}
\end{figure}

\begin{figure}
\includegraphics[ width=0.49\textwidth,height=0.125\textwidth]{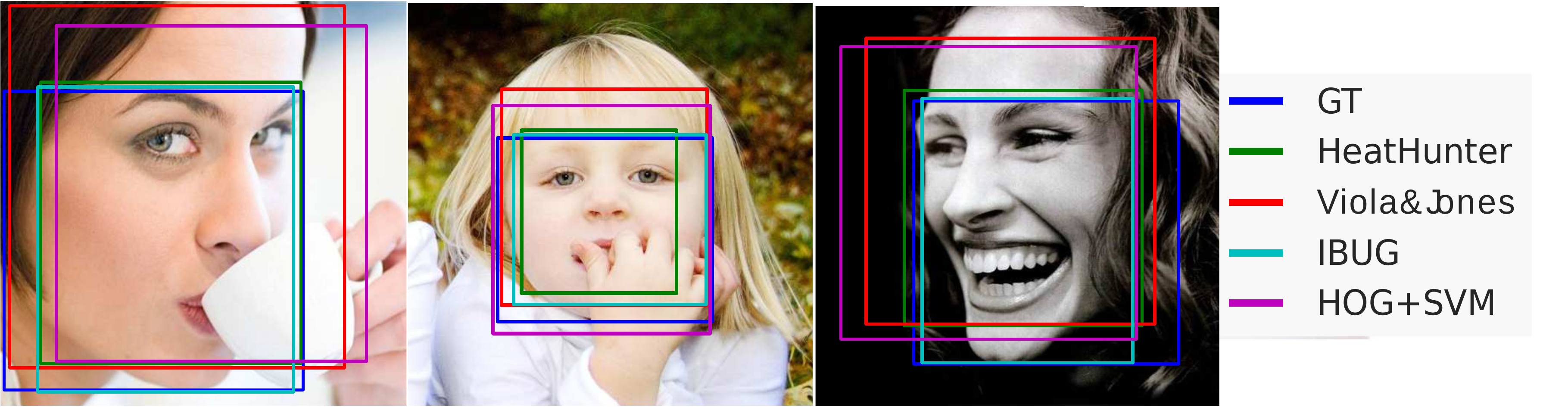}\\
\caption{Different face detections on example images.}
\label{fig::facebbexample}
\end{figure}

None of these detectors guarantees 100\% detection rate due to the difficulty of those face samples. Taking the IBUG bounding box as ground truth, the missing rate for V\&J, HeadHunter and HOG+SVM are 16\%, 5\% and 8\% respectively. Some papers only make  comparison on the successfully detected faces, which make further comparison hard as those samples missed by the detectors are usually difficult and very influential to the overall performance. Some methods also use a global bias to adjust different face detections \cite{mathias2014face}, however, as shown in Fig~\ref{fig::facebbexample} the bias between different face detections are face image dependent thus it is inappropriate to use a single bias for adjustment. Therefore, we propose a deep convolutional network (ConvNet) regression approach to fulfil the missing detections. More specifically, we select reliable samples to train a ConvNet that takes the image content from the IBUG bounding box as input and predicts the bounding box differences. The samples with successful detections in the training set and their mirrored samples are used for training the ConvNet model. The process and the ConvNet structure are illustrated in Fig.~\ref{fig::facebbnet}. After each convolution and fully connected layer (except the final one), ReLU non-linear operation is applied. In this way, for each training and testing sample, we obtain three types of reliable face detections (V\&J, HeadHunter, HOG+SVM), in addition to the IBUG bounding boxes. 

We test the effectiveness of the ConvNet using the samples in the testing set with successful detections by the original detectors. The results are shown in Table \ref{tab::facedetectionrate}. Almost all samples are with IOU $>$ 0.5 (IOU: Intersection Over Union). We checked the samples with IOU $<$ 0.5 and found that it was caused by  incompleteness of a  few faces. 
We  obtain high average IOU for all the three detectors and believe that the face detections fulfilled by our ConvNet regression are accurate and reliable. 

\begin{table}[h]
\centering
\setlength{\tabcolsep}{2pt}
\footnotesize
\caption{Face regression by ConvNet.}
\begin{tabular}{lccc}
\hline
Face detection     & Viola\& Jones  & HeatHunter & HOG+SVM \\
IOU $>$ 0.5 (\%)    & 99.99  &        100    &   99.99       \\
Average IOU & 0.92 & 0.90 & 0.89 \\
\hline
\end{tabular}
\label{tab::facedetectionrate}
\end{table}

\begin{figure}
\centering
\includegraphics[width=0.49\textwidth,height=0.16\textwidth]{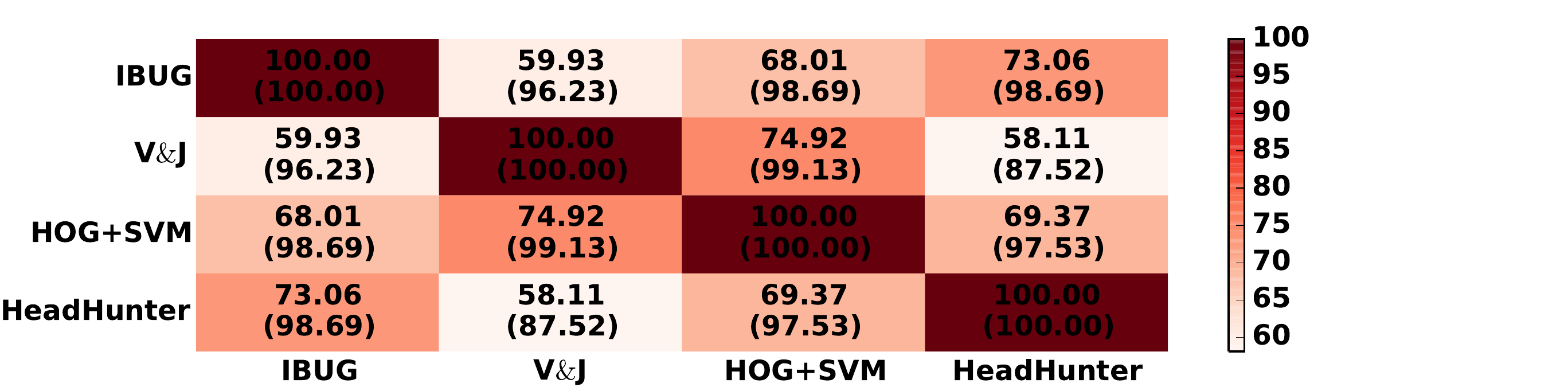}\\
\caption{Different face bounding boxes in 300W++. Average IOU (upper) and percentage (lower) of IOU $>$ 0.5. }
\label{fig::facebbvsfacebb}
\end{figure}

The statistics of different face bounding boxes of 300W++ are shown in Fig. \ref{fig::facebbvsfacebb}. According to the conventional face detection measure (IOU), most cross-method bounding boxes can be regarded as \textit{correct} detections since their IOU values are usually bigger than 0.5. However, as we will show in the experimental section, such face detection variance will lead to very different face alignment results.

\subsection{A new evaluation metric: AUC$_{\alpha}$}

Similar to other methods in the literature, landmark-wise face alignment error is first normalised in the following way to make it scale invariant:
\begin{equation}
e_{\mathrm{x}} = \frac{||\mathrm{\hat{x}} - \mathrm{x}^{GT}||}{D_{IOD}}
\end{equation}
where $||\mathrm{\hat{x}} -\mathrm{ x}^{GT}||$ is the Euclidean distance between the estimated location $\mathrm{\hat{x}}$ and the true location $\mathrm{x}^{GT}$. $D_{IOD}$ is the inter-ocular distance, i.e. Euclidean distance between two eye centres.

In recent years, with the rapid progress of face alignment, most of the recent approaches  report a error level of $e$ at around 0.05 or smaller, that is close to human performance. This is a proper error metric for one landmark or one face image. In order to evaluate the performance of a set of images, there are mainly two methods. One is the mean error, sample-wise, landmark-wise or overall. The other one is the Cumulative Error Distribution (CED) curve, which is the cumulative distribution function of the normalized error. Using mean error to measure the performance is very straightforward and intuitive given its single value form. However, this measure is heavily impacted by the presence of some big failures, i.e., outliers, in particular when the average error level is very low. In other words, the mean error measure is very fragile even if there are just a few images with big errors. For example, given a set of 500 evaluation samples, if there are 5 failures with mean error at 1.0 and the rest of them are with average mean error at 0.03, then the overall mean error increases from 0.03 to 0.04, which is equivalent to 33\% of increase. Thus we argue that though the mean error is widely used for face alignment evaluation like \cite{jourabloo2015iccv,antonakos2015active,lee2015face,renface,wu2014hierarchical,zhang2014facial,burgos2013robust,wu2013facial,cnnficalcvpr2013}, it does not  provide a big picture on which  cases the errors occur (minor big alignment error or many inaccuracies). We will further validate this argument in Section \ref{sec::BC_offtheshelf}. 

\begin{figure}
\centering
\includegraphics[ width=0.47\textwidth,height=0.32\textwidth]{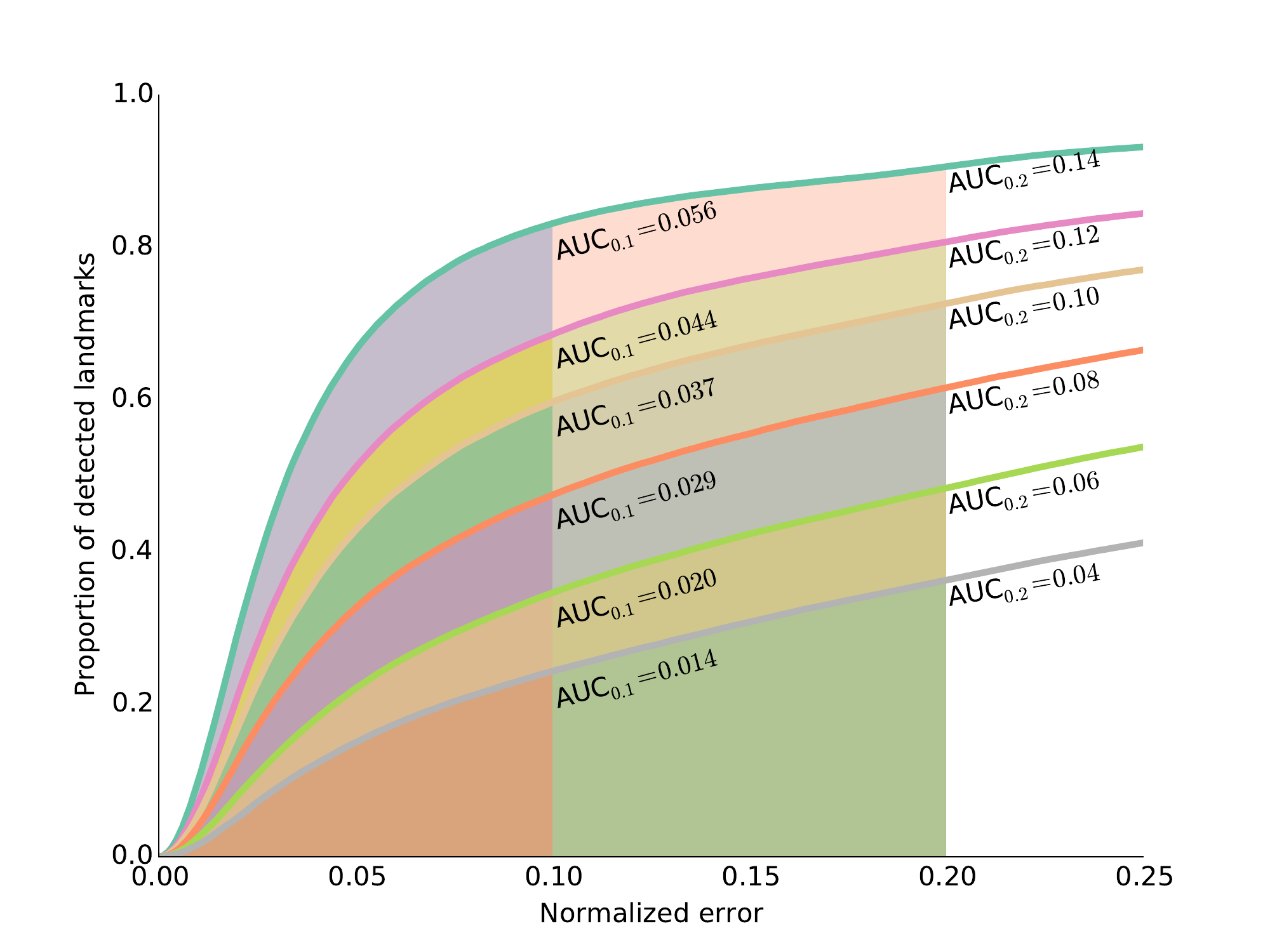}\\
\caption{The evaluation metric AUC$_{\alpha}$ with various values given $\alpha = 0.1$ and $\alpha  = 0.2$.}
\label{fig::auc}
\end{figure}

In terms of outliers handling, CED is a better way. However, it is not intuitive given its curve representation.  It is also hard to use it in sensitivity analysis. Therefore, we propose a new evaluation metric called AUC$_{\alpha}$. It is defined as:
\begin{equation}
\label{eq::errormetric}
AUC_{\alpha} = \int_0^{\alpha}f(e)de 
\end{equation}
where $e$ is the normalized error, $f(e)$ is the cumulative error distribution (CED) function and $\alpha$ is the upper bound that is used to calculate the definite integration. Given the definition of the CED function, the value is AUC$_{\alpha}$ lies in the range of $[0, \alpha]$. The value of AUC$_{\alpha}$ will not be influenced by points with error bigger than $\alpha$. In Fig.~\ref{fig::auc}, we plot the several example values of AUC$_{\alpha}$ and their corresponding CED curves to give readers an idea of our proposed metric. In the rest of the paper, we will use AUC$_{\alpha}$ as the main metric for performance evaluation.

\section{Experiments}
\subsection{Basic comparison of off-the-shelf models}
\label{sec::BC_offtheshelf}
We first make a comparison of the state of the art methods based on their off-the-shelf models, i.e., the models provided by the authors.  
These models might trained in different settings (e.g. the use of face detection, the number of facial landmarks and training samples) and some of them are lack of detailed descriptions.  It makes fair comparison of the model performance difficult. However, it is still valuable to evaluate their relative merits on the same testing data and carry out sensitivity analysis of these models w.r.t face detection variation. We first identify the \textit{best} face detection for each model either by the information provided by the original paper, or by comparing their performance or by communication with the authors. The related information of each model is shown in Table. ~\ref{tab::all_ots_methods}. We note that the trained model and the code of LBF \footnote{LBF: https://github.com/jwyang/face-alignment} and RCPR \footnote{RCPR: https://github.com/ChrisYang/RCPR} is from open source implementation, not from the original authors. The rest of the models are all from the original authors. We also record the run-time performance in the table. 
 Since the methods are implemented in different languages, the comparison here might not be very fair. Most of the methods have comparable speed to the original description, except the LBF, possibly due to the re-implementation in Matlab. 
As most of the models are trained to detect 68 facial landmarks and some of them detect a sub-set of them, we record and calculate the landmark-wise error of all detected landmarks, instead of using their common 49 landmarks. This gives a slight advantage to the methods that detect only the inner points as the landmarks along the face contour are generally regarded more difficult. We do so in order to keep consistent with the baseline dataset, also to make future comparison more convenient. 

\begin{figure*}
\centering
\includegraphics[trim={1cm 0 1cm 1cm},clip,width=0.45\textwidth,height=0.25\textwidth]{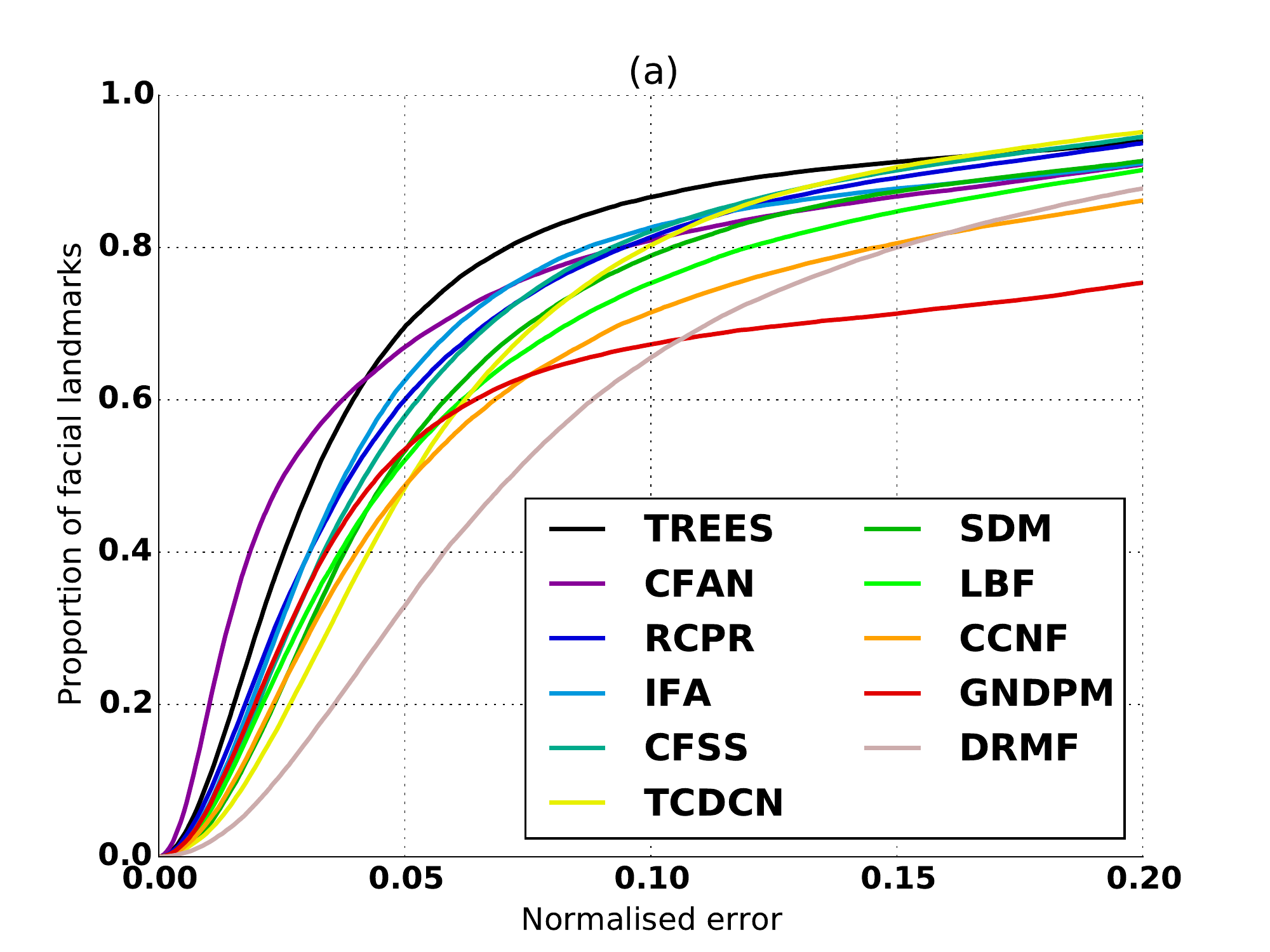}
\includegraphics[trim={1cm 0 1cm 1cm},width=0.45\textwidth,height=0.25\textwidth]{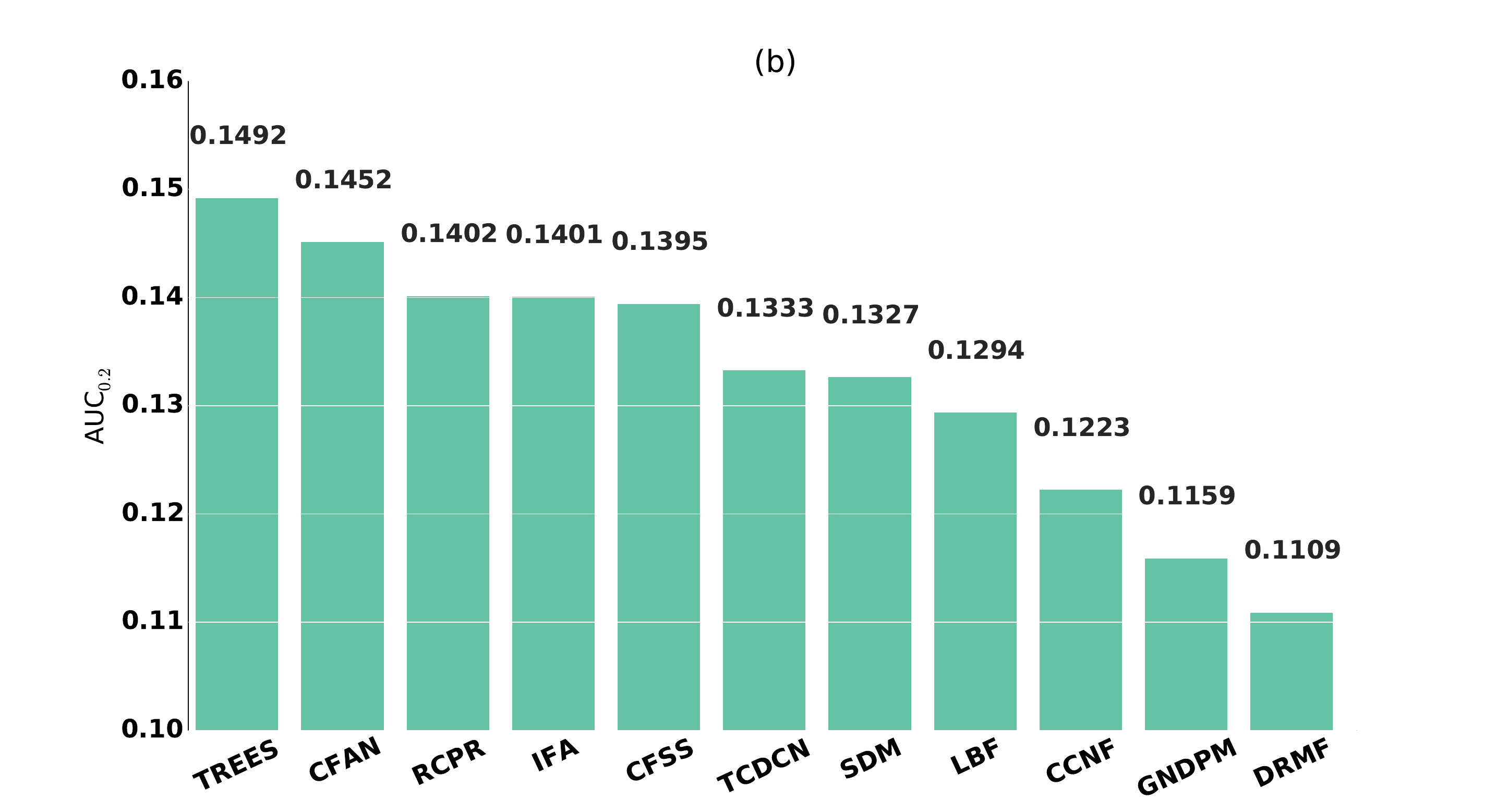}\\
\includegraphics[trim={1cm 0 1cm 0cm},width=0.45\textwidth,height=0.25\textwidth]{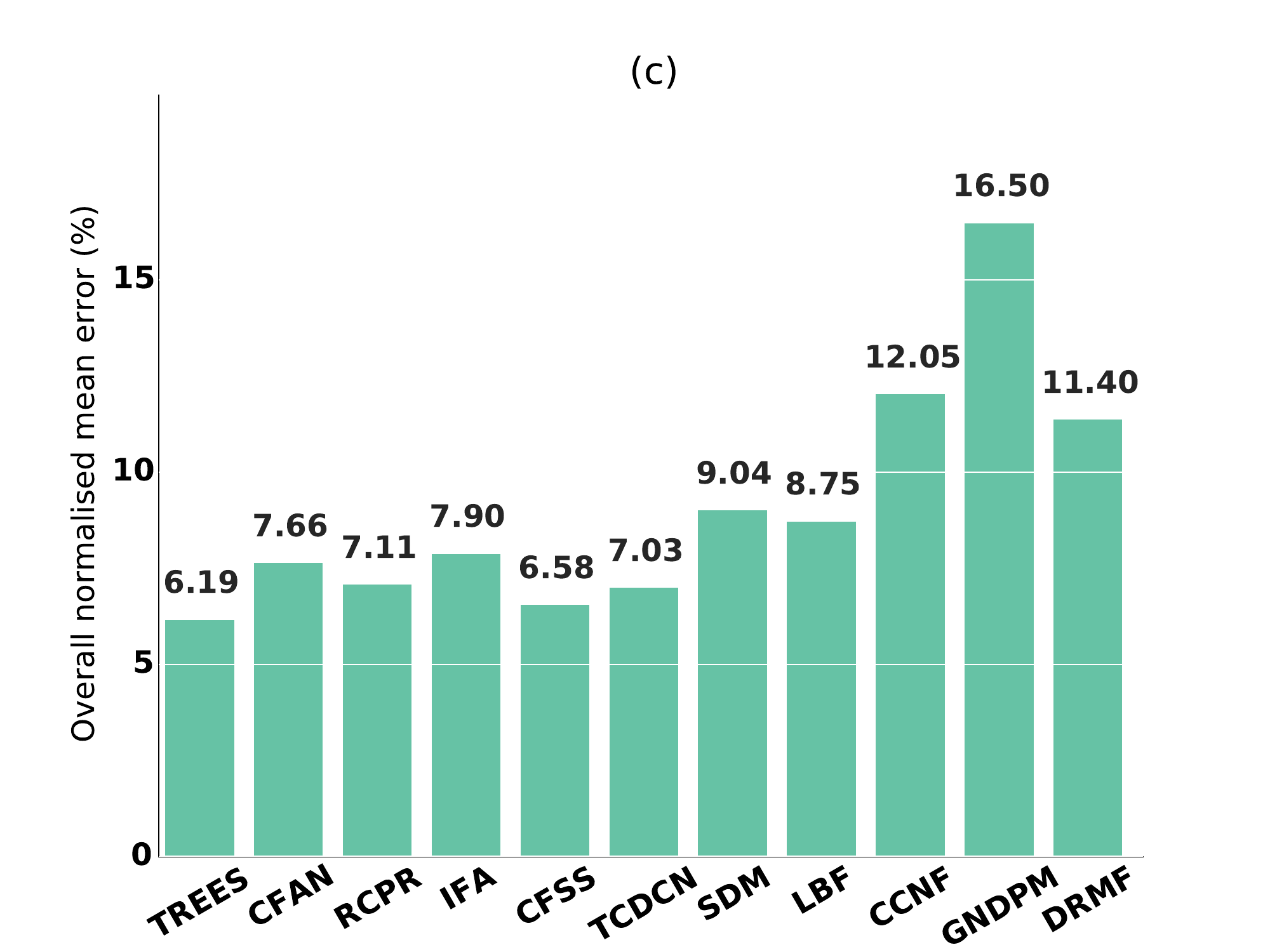}
\includegraphics[trim={1cm 0 1cm 0cm},width=0.45\textwidth,height=0.25\textwidth]{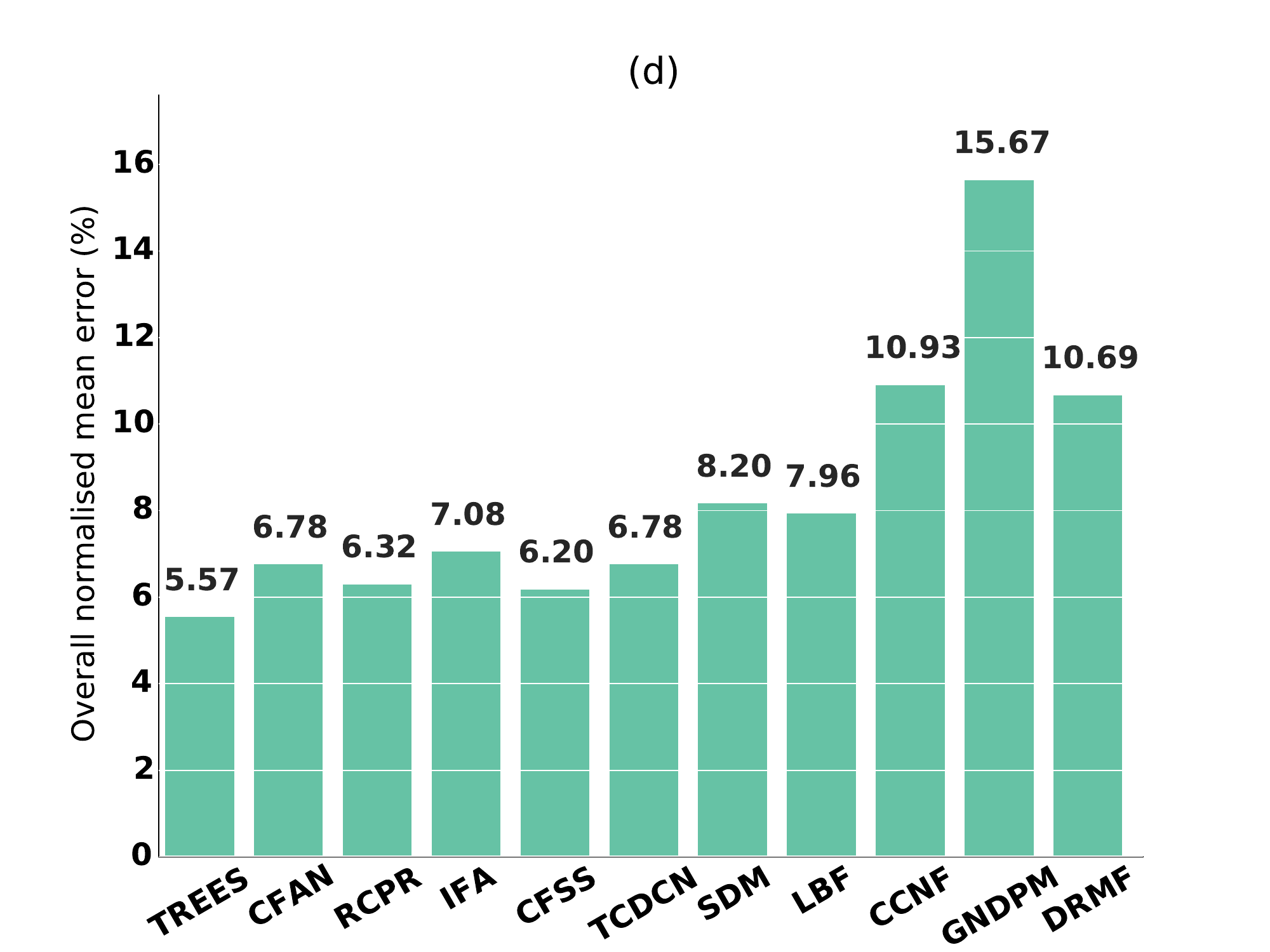}
\caption{Performance comparison of different methods based on the best face detectors. (a) cumulative error distribution. (b) plot of AUC$_{0.2}$. (c) Mean error on all test images. (d) Mean error on test set excluding the top 5 erroneous images. }
\label{fig::facebestbb}
\end{figure*}

We first plot the CED in Fig.~\ref{fig::facebestbb} (a). As can be seen, for methods that have very close performance, it is not intuitive to evaluate them using the curve plots. We plot our proposed AUC$_{\alpha}$ (with $\alpha=0.2$) in Fig.~\ref{fig::facebestbb} (b), it is very straightforward to illustrate the performance of different methods. TREES model performs the best, followed by CFAN, and three models (RCPR, IFA, CFSS) have very similar performance. We also plot the overall mean error in Fig.~\ref{fig::facebestbb} (c). As can be seen, the order is not consistent with the reverse order of AUC$_{\alpha}$. For example, the mean error of RCPR, CFSS and TCDCN is lower than CFAN, however, from the curve plot, we can observe CFAN performs better than them when the error level is below 0.1, based on fact that a detection is usually regarded as successful when error is smaller than 0.1. CFAN and IFA have the similar mean error value (7.66 vs. 7.90) but their performance is quite different from each other. We further conduct an experiment to show that the mean error can be very easily influenced by big error samples, even a small number of them. More specifically, for each method, we excluded 5 top erroneous samples (they are all failures to the method) out of 689  and then re-calculated the mean error. The result is plotted in Fig.~\ref{fig::facebestbb} (d). As can be seen, despite only 5 erroneous samples are excluded (less than 1\%), the value of mean error changes very significantly, 0.73 in average, which is even much bigger than the difference between several methods (e.g. 0.55  for CFAN vs. RCPR, 0.45 for TCDCN vs. CFSS, 0.29 for SDM vs. LBF). It validates our argument that mean error is not a effective evaluation measure: 1) it sometimes does not reveal the actual relative merits; 2) it is very sensitive to big alignment error, especially on dataset like 300W with some challenging samples that might result in big alignment errors. On the contrary, our proposed AUC$_{\alpha}$ is a more effective and consistent metric. 


\begin{table*}[!hbtp]
\footnotesize
\setlength{\tabcolsep}{3pt}
\centering
\caption{Evaluated face alignment methods and their properties. (NC: Not Clear)}
\label{tab::all_ots_methods}
\begin{tabular}{lccccccccccc}
    \hline
    Methods                   & SDM             & RCPR            & IFA     & LBF  & TREES   & CFAN & TCDCN   & GNDPM & DRMF            & CCNF & CFSS      \\
    Best BB                   & V\& J & V\& J & HOG+SVM & IBUG & HOG+SVM & IBUG & IBUG & IBUG  & V\& J & IBUG &IBUG     \\
   Landmarks \# & 49              & 68              & 49      & 68   & 68      & 68   & 68      & 49    & 66              & 68        &68\\
    Training set              & NC     & 300W            & 300W    & 300W & NC    & NC & CelebA + 300W   & 300W  & NC            & 300W+MPIE &300W\\
    Run-time (FPS)                 & 40              & 80              & 20      &10  & 300     & 20   & 50      & 70    & 0.5             & 30        &  10\\ \hline
\end{tabular}
\end{table*}

\subsection{Sensitivity analysis of off-the-shelf models}
\subsubsection{Face centre shifts}
\begin{figure}[!hbt]
\centering
\begin{subfigure}[t]{0.2\textwidth}
\centering
\includegraphics[height=0.98\textwidth]{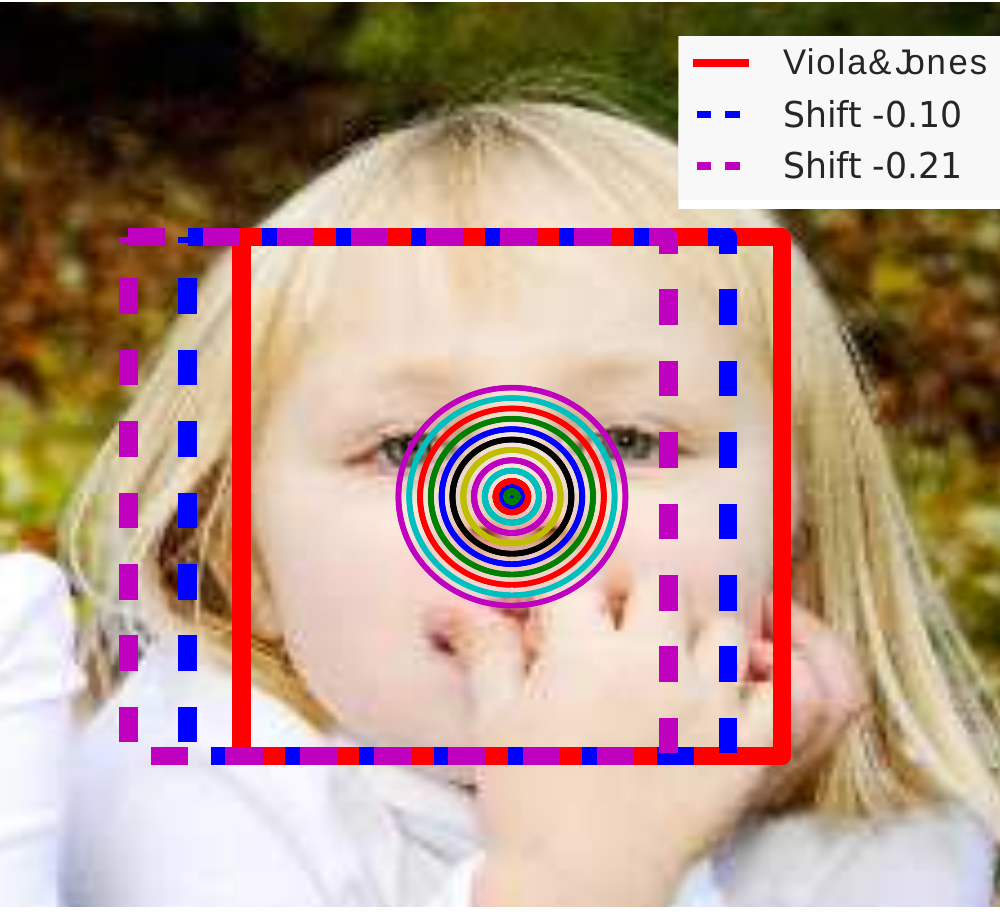}
\caption{}
\end{subfigure}
\begin{subfigure}[t]{0.27\textwidth}
\centering
\includegraphics[height=0.9\textwidth]{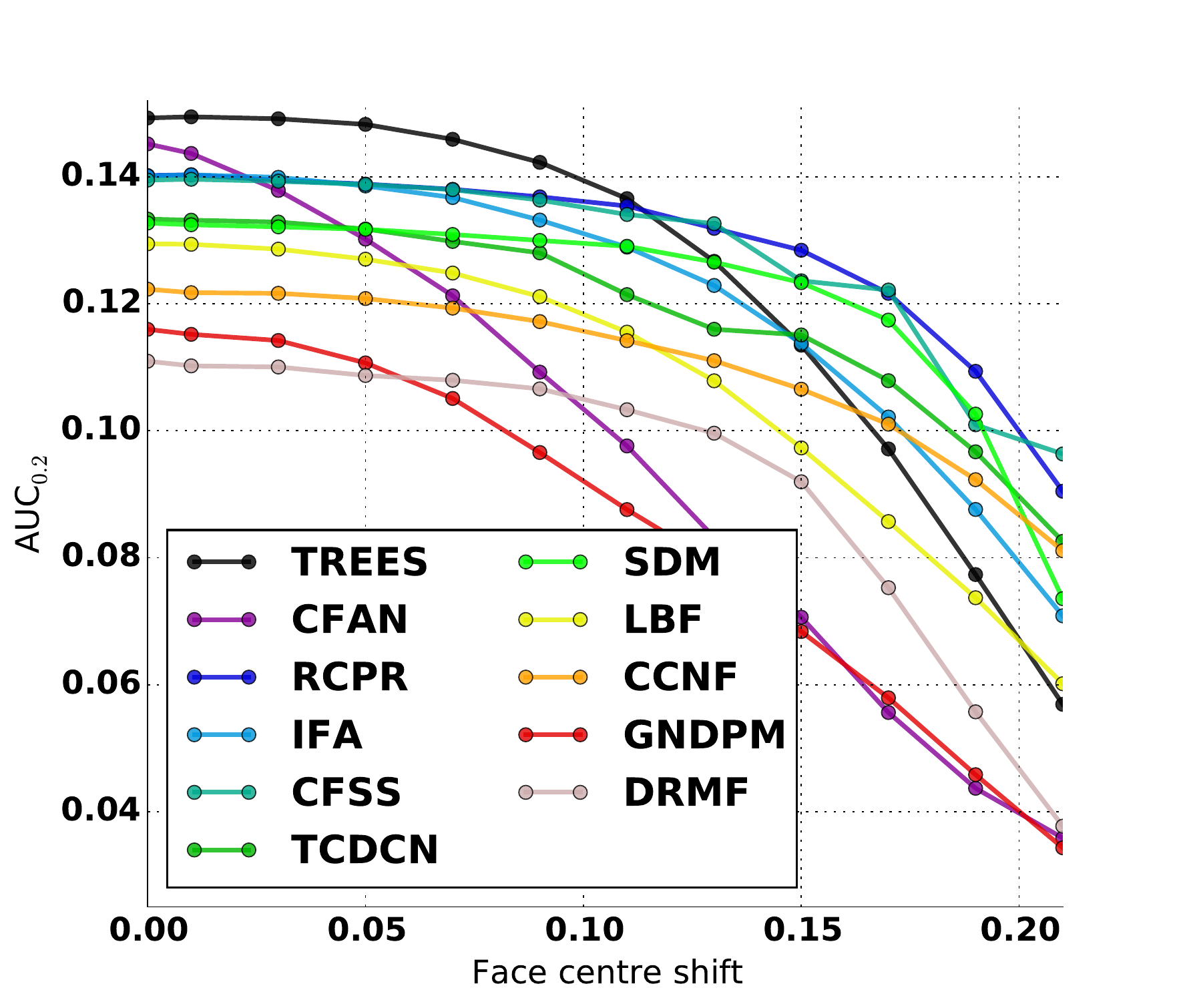}
\caption{}
\end{subfigure}
\caption{Experiments on synthesised face centre shifts. (a) Face centre shift synthesis. (b) AUC$_{0.2}$ values vs. face centre shifts.}
\label{fig::facecentershift}
\end{figure}

We first evaluate the impact of face centre shifts on the face alignment. Given a face bounding box, we make random permutation of the bounding box centre. More specifically, we set a range of radius on how much the centre shifts from the original location. The radius are  $[0.01:0.02:0.21]$ of the face bounding box scale (mean value of width and hight), which is demonstrated by the circles in different colours in Fig.~\ref{fig::facecentershift} (11 radius in total). We keep the face bounding box size unchanged. For each radius, we randomly select 10 locations on the arc as the new face centre to get 10 different bounding boxes. We note that for a given method, the shift is carried out on its \textit{best} face bounding box described in Table~\ref{tab::all_ots_methods}, and for the methods that share the same \textit{best} face bounding box, we use the same set of randomly generated bounding boxes for a fair comparison. The new bounding boxes are fed to the face alignment models, together with the image content to get the alignment output. In this way, for each radius, we get 10 groups of detection results. Then we calculate the overall performance using AUC$_{0.2}$. For each radius, we calculate a mean value from 10 runs. Thus for the 11 evaluated methods, we ran 11x11x10 groups of experiments for sensitivity analysis.  The comparison is shown in Fig.~\ref{fig::facecentershift} (b). The performance of all the evaluated methods drops while the face  shifts. Some methods like SDM, RCPR, CFSS, CCNF are robust than others to small range of face centre shits (less than 5\% of the face scale, which is very common in real scenarios). There are some methods, e.g. CFAN, is very sensitive to even a small amount of face centre shifts. 
AUC$_{0.2}$ value drops from 0.145 (original) to 0.137 (3\% shift), to 0.130 (5\% shift) and to 0.121 (7\% shifts). With 10\% of face shift, its performance drops from second best to second worst.  One can refer to Fig.~\ref{fig::auc} with curve plots on various AUC$_{\alpha}$ values to get an idea about how much such change represents. GNDPM is the second sensitive model while TREES model becomes very sensitive w.r.t shift bigger than 10\%.  


\subsubsection{Face scale changes}
\begin{figure}[!hbt]
\centering
\begin{subfigure}[t]{0.2\textwidth}
\centering
\includegraphics[height=0.9\textwidth]{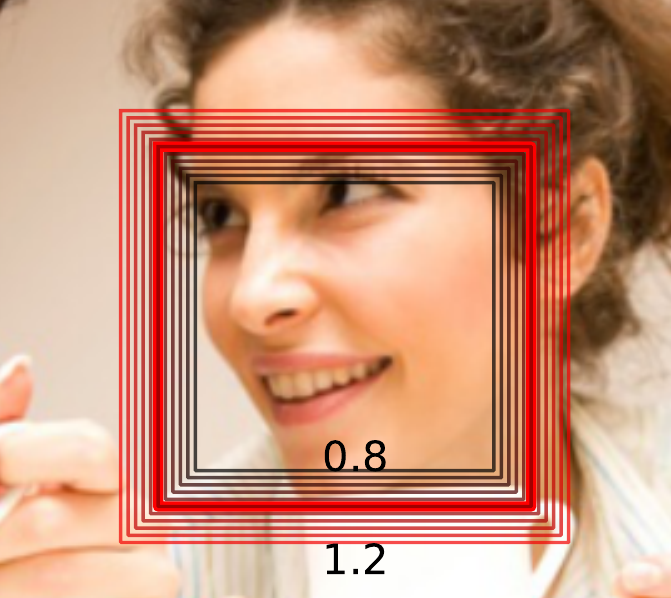}
\caption{}
\end{subfigure}
\begin{subfigure}[t]{0.27\textwidth}
\centering
\includegraphics[height=0.87\textwidth]{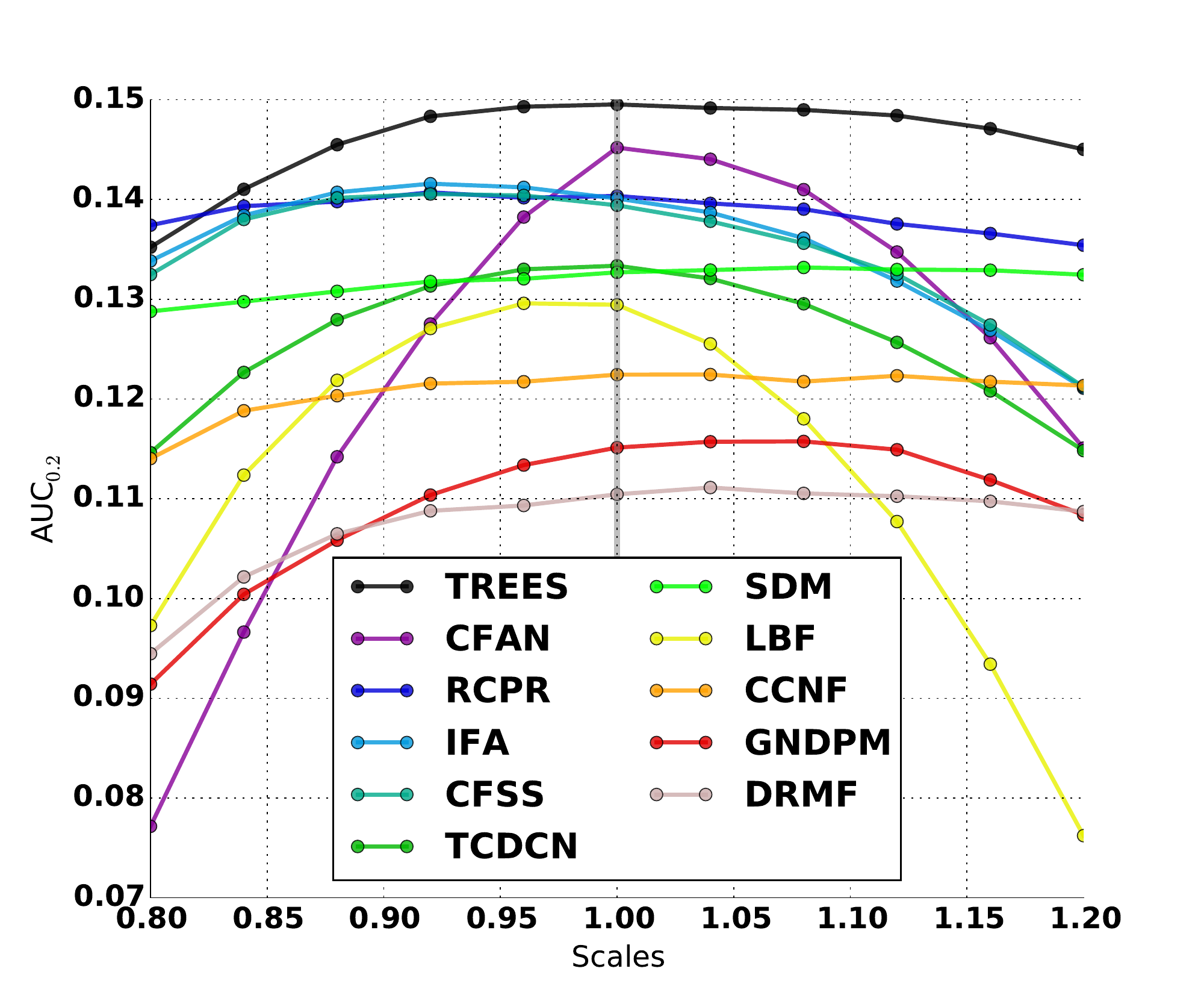}
\caption{}
\end{subfigure}
\caption{Experiments on synthesised face scale changes. (a) Face scaling, with original face bounding box highlighted by red shadow. (b) AUC$_{0.2}$ vs. face scale.}
\label{fig::facecescale_example}
\end{figure}

We then evaluate the impact of face scale changes on the face alignment. We re-scale a face bounding box by a ratio that ranges from 0.8 to 1.2, as shown in Fig.~\ref{fig::facecescale_example} (a). Again, the rescaling is carried out on the \textit{best} face bounding box for each method. Compared to the results on face centre shifts, most of the evaluated models demonstrate better robustness against scale changes, especially the best performing methods, like TREES, RCPR and SDM. However, similar to situation on face centre shifts, CFAN is the most sensitive methods. The AUC$_{\alpha}$ against the decreasing scales or increasing scales forms a steep line. For example, when the face scale is 10\% smaller (the IOU is still ~0.8), $AUC_{0.2}$ drops from 0.145 to 0.127. LBF model is also very sensitive to scale changes, especially when scale increases. RCPR, CFSS and IFA has similar original performance but RCPR shows better robustness vs. scale changes.

\subsubsection{Experiments on real face detection shifts}
In this section, we evaluate the face alignment methods on various face detection results from different real face detectors.

\begin{figure}
\centering
\includegraphics[trim =0.0cm .0cm 3.0cm .0cm, clip = true, width=0.5\textwidth,height=0.17\textwidth]{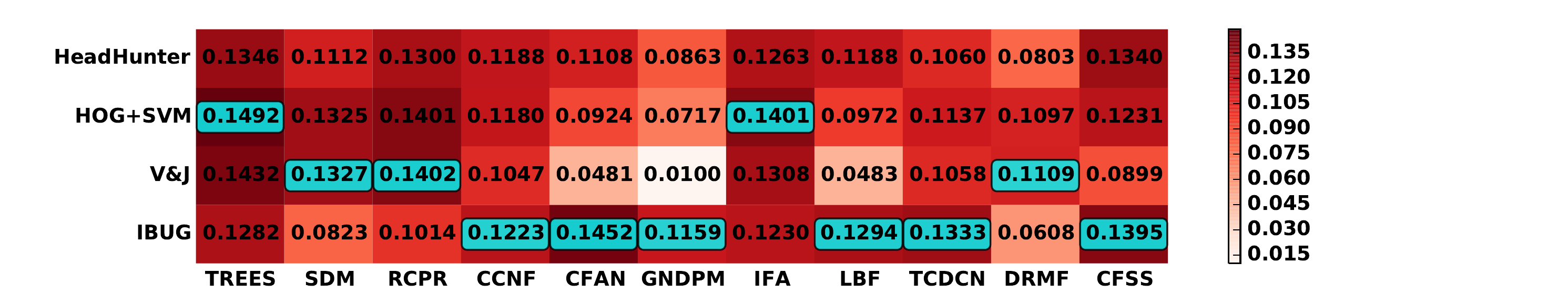}\\
\caption{Face alignment performance (AUC$_{0.2}$) of different models on various face detection results. The result from \textit{best} face box is highlighted in light blue.}
\label{fig::realfaceshift}
\end{figure}

The results are shown in Fig.~\ref{fig::realfaceshift}. As can be seen, when the face detection is switched away from the \textit{best} detector, the performance deteriorates in certain amount. The results from similar face bounding boxes (like HOG+SVM and Viola-Jones, with 75\% average IOU) are close for most of the methods like TREES, SDM, IFA and DRMF. But for sensitive methods like CFAN and LBF, the results from similar face bounding boxes are still very different (0.0924 vs. 0.0482 for CFAN  and 0.0972 vs. 0.0483 for LBF). In the extreme case, like the model of GNDPM, when the face detector changes from \textit{best} IBUG to Viola-Jones, it hardly gives any reliable result. The models trained with IBUG show better results on HeadHunter than on HOG+SVM and Viola-Jones due to their similar bounding box definition. In general, none of the evaluated method is able to keep its performance when face detection shifts from its \textit{best} face detector, despite that fact that most of them have high overlap ratio according only from the perspective of face detection. HeadHunter is one of the best performing face detectors in the literature, nevertheless, since none of the face alignment models are trained with it, their performance degrades, especially for those trained with Viola-Jones detectors like SDM (0.1327 vs. 0.1113), RCPR (0.1402 vs. 0.1300) and DRMF (0.1110 vs. 0.0803). Therefore, it is essential to keep the same face detector in training and testing stage, for either performance evaluation or practical application.


\subsection{Sensitivity analysis of re-trained models}
\begin{figure}
\centering
\includegraphics[trim =0.0cm .0cm 0.0cm .0cm, clip = true, width=0.23\textwidth,height=0.18\textwidth]{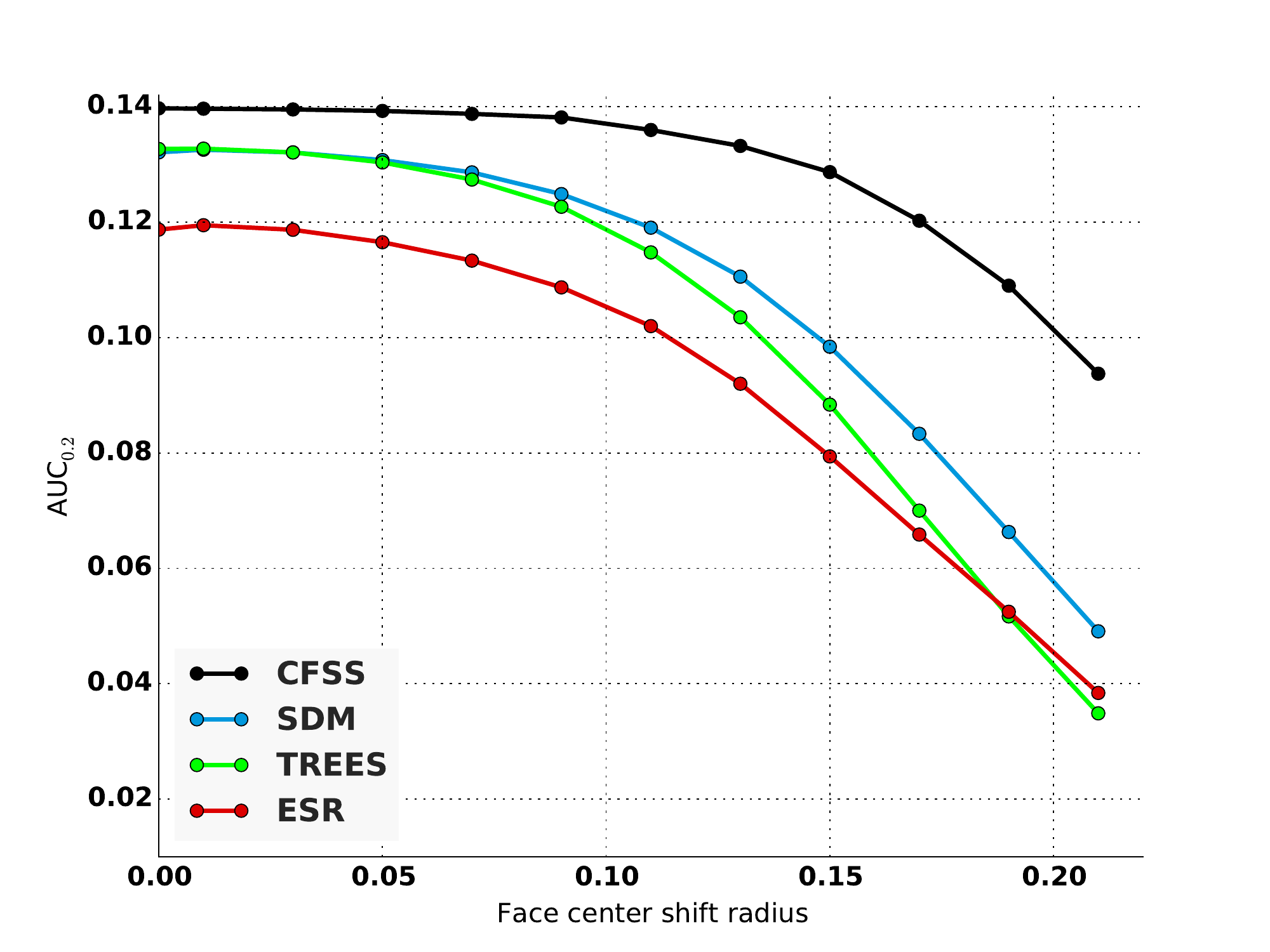}
\includegraphics[trim =0.0cm .0cm 0.0cm .0cm, clip = true, width=0.23\textwidth,height=0.18\textwidth]{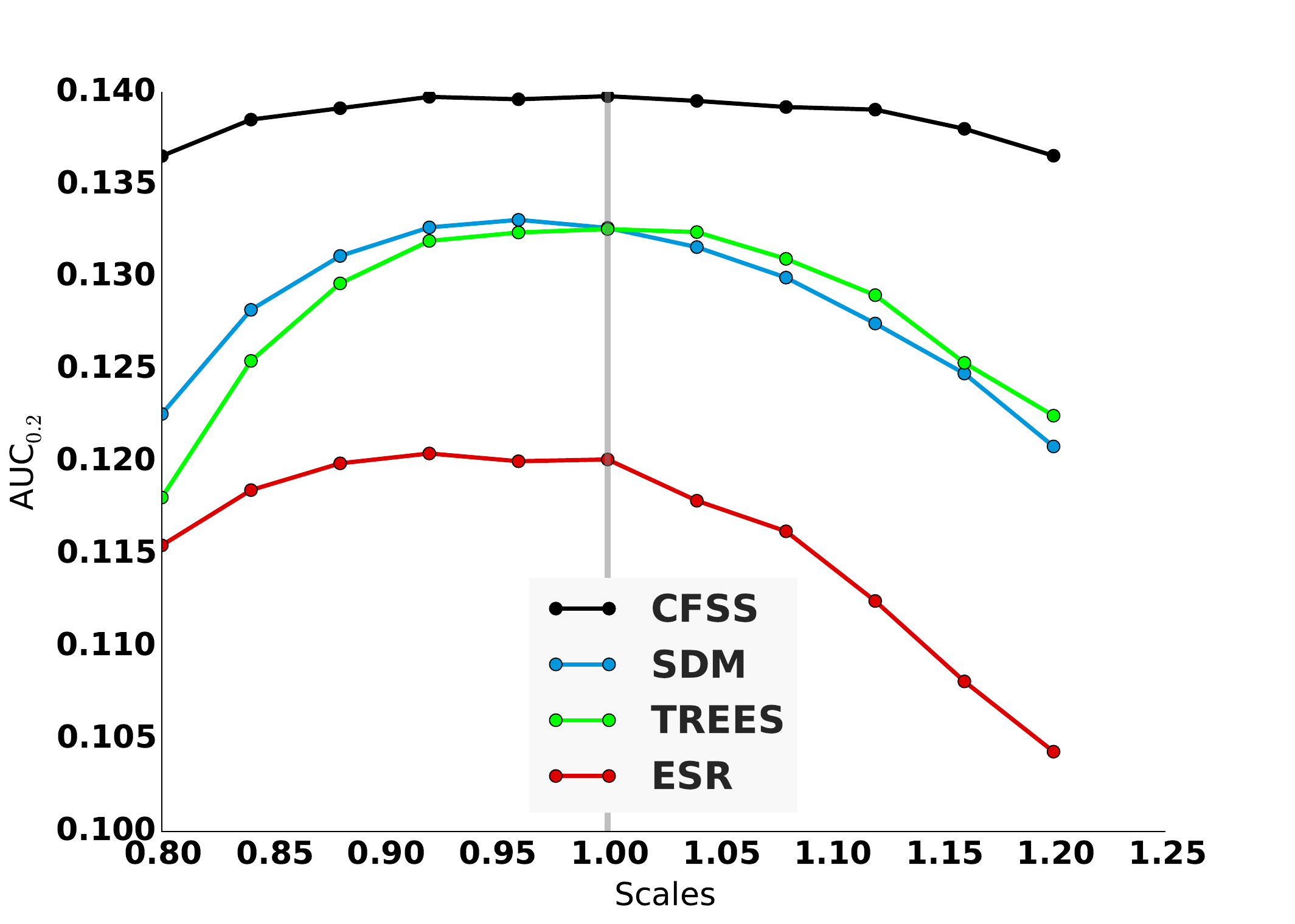}
\caption{Sensitivity w.r.t centre shift (Left) and face scale (Right) of re-trained models.}
\label{fig::retrainface}
\end{figure}

By using the off-the-shelf model, we are unable to make a fair comparison cross different methods due to the difference in experimental setting, training data, and face detection. We select four representative methods for a  fairer comparison by re-training their models using their default setting on the same training data and the same data setting. They are CFSS \cite{zhu2015face}, TREES~\cite{kazemi2014one}, SDM \cite{xiong2013supervised} (re-implemented by \cite{zhu2015face}) and ESR \cite{suncvpr2012}. The first two methods show very good performance in our previous comparison. The later two are the most popular methods in recent years. We re-train the models on 300W++ training set using the HOG+SVM face detection. We chose this as it is a very good trade-off between effectiveness and efficiency. As all of them are cascaded methods, we set the random initialisation number to 20. We carry out sensitivity analysis in a similar way as we did before. The sensitivity against face centre shift and face scale changes are demonstrated in Fig.~\ref{fig::retrainface}.

As can be seen, CFSS shows the best performance in both localisation accuracy and robustness against the initialisation variation. TREES and SDM has very close performance, better than the ESR. Similar to what we found in previous experiment, face centre shift has high influence on the performance while  that of scale change is lower.     

By comparing to the performance of the off-the-shelf models in Fig.\ref{fig::facebestbb}, we can further observe that 1) the training setting has very significant impact on performance as well, e.g., for TREES, the AUC$_{0.2}$ drops from 0.149 to 0.123 due to different setting and possibly training data difference; 2) the relative ranking order of CFSS and TREES changes as CFSS surpasses TREES in AUC$_{0.2}$ under the same setting. However, the computational expense and model complexity of CFSS is much higher than TREES as shown in Table \ref{tab::all_ots_methods}. 

\subsection{Important factors}
We study the important factors in holistic-based methods. We choose Explicit Shape Regression (ESR), as it is regarded as a breakthrough face alignment method in both accuracy and efficiency and widely adapted ever since. Two factors, namely initialisation method and cascade level, are studied below.
\subsubsection{Initialisation}
\begin{figure}[!hbt]
\centering
\begin{subfigure}[t]{0.23\textwidth}
\centering
\includegraphics[height=0.8\textwidth]{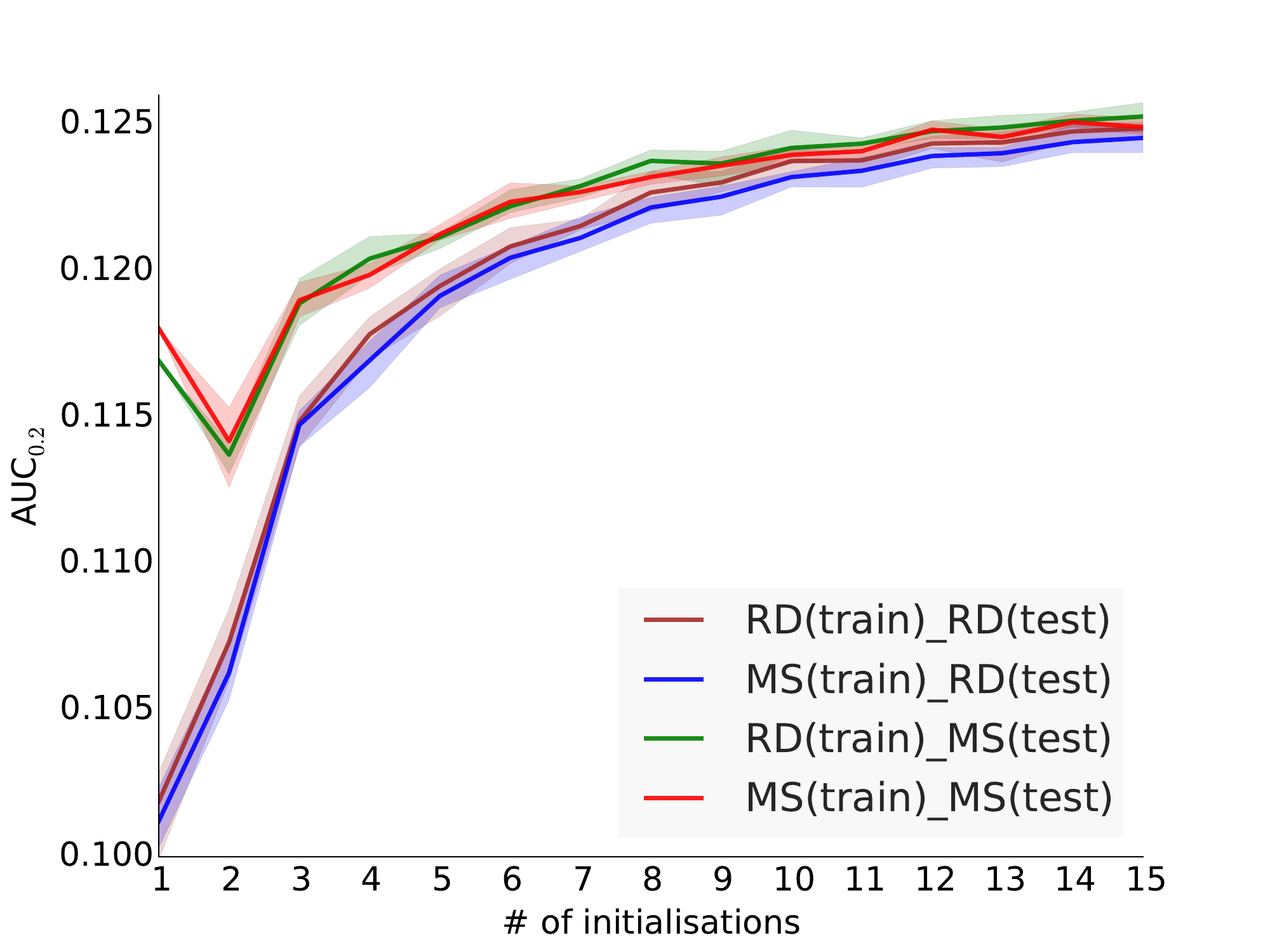}
\caption{Initialisation}
\label{fig::importantfactor_init}
\end{subfigure}
\begin{subfigure}[t]{0.24\textwidth}
\centering
\includegraphics[height=0.8\textwidth]{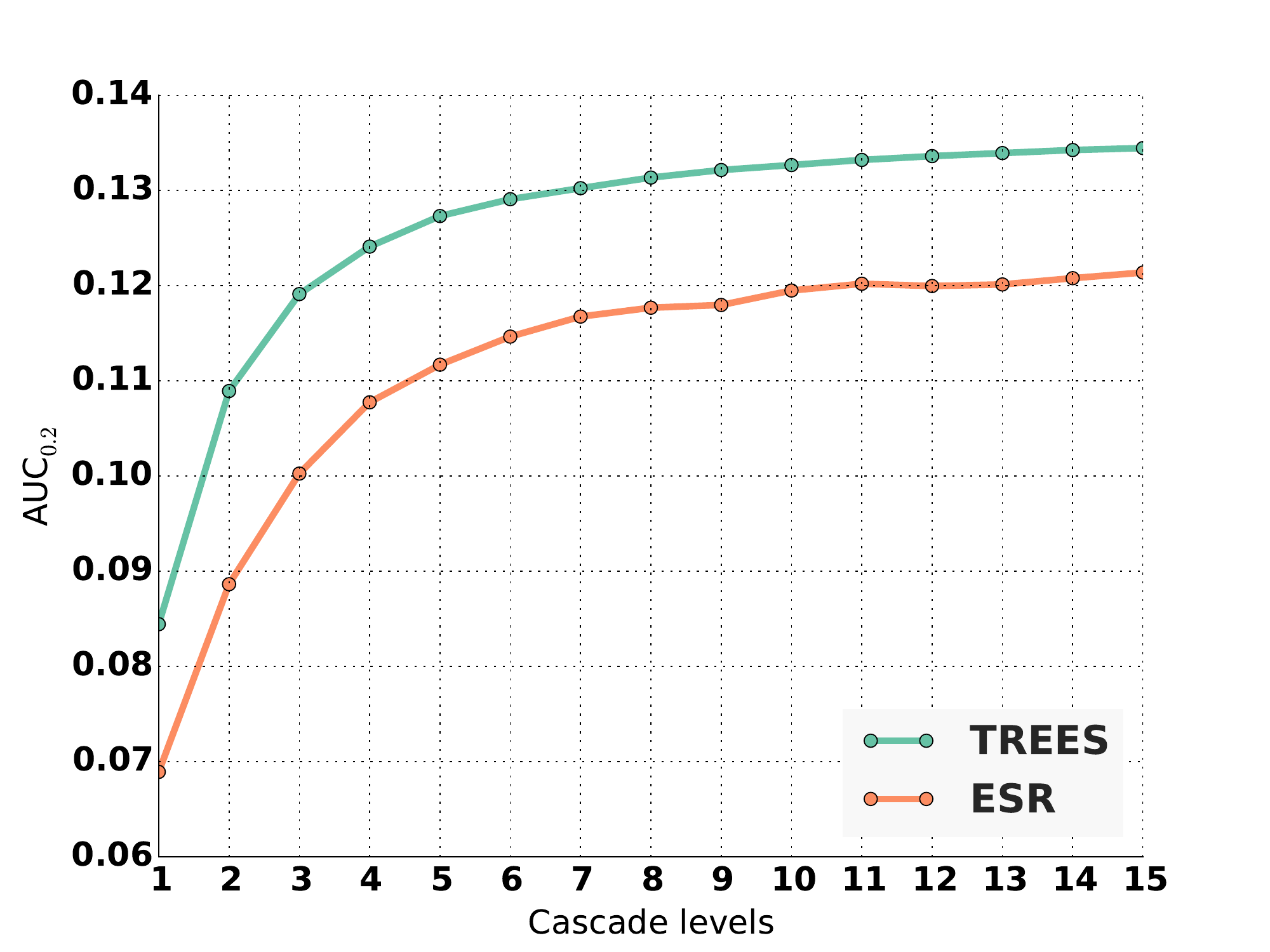}
\caption{Cascade level}
\label{fig::importantfactor_Cascade}
\end{subfigure}
\caption{Important factors evaluation.}
\end{figure}

ESR proposed a multiple-random-initialisation scheme for performance boost. As we discussed in Sec.\ref{sec::methods}, another (unsupervised) initialisation scheme is using Mean Shape (MS). We carry out experiments to study how different they are in practice. We trained two models, with the same augmentation number (20) and one (\textbf{RD}) with all RD initialisations, the other (\textbf{MS}) with 1 MS + 19 RD initialisations . At testing stage, we recorded the results of using $N$ initialisations (final result is calculated as the median value) in two schemes, one (\textbf{RD}) with $N$ RD  and the other (\textbf{MS}) with 1 MS and $N$-1 RD. Thus we get 4 combinations: RD(train)-RD(test), RD(train)-MS(test), MS(train)-RD(test), MS(train)-MS(test). Since randomness was involved in this process, we repeated this experiment for 5 times. The results for $N = \{1,...,15\}$ are shown in Fig.~\ref{fig::importantfactor_init}. We can draw three useful conclusions from the results: 1) using MS in testing is always useful, regardless how the model is trained (RD or MS); 2) increasing the number of initialisations leads to non-decreasing performance; 3) using MS during training time has little impact on the performance. Therefore in practical applications we suggest to always use MS as one of the initialisations, given the fact that using just one MS has similar performance to that from 4 RDs while being 4 times faster.

\subsubsection{Cascade levels}
We further study the impact of cascade levels. We first train model with deep cascade. Then we test the performance of the model with various levels of cascade. The results for ESR and TREES are shown in Fig.~\ref{fig::importantfactor_Cascade}. As can be seen, deeper cascade can lead to non-decreasing performance boost, at a cost of bigger model size and longer run-time. This is opposed to the finding of feature based AAMs \cite{antonakos2015feature}, that deeper iteration might lead to over-fitting. For practical application, using 10 cascade levels will lead to a reasonable performance for both ESR and TREES. 


\section{Conclusion}
In this paper, we have presented our empirical study on recent face alignment methods. We first extended the 300W dataset and formed the 300W++ dataset with more practical face detections. We then proposed a new face alignment evaluation criterion AUC$_{\alpha}$ that is very effective in measuring the performance with a single value. Based on this, we carried out sensitivity analysis and comparative study of several representative face alignment methods including their off-the-shelf models and re-trained models. We also studied several influential aspects in cascaded face alignment. From a comprehensive empirical study, we drew useful conclusions of current face alignment methods and made  insightful suggestions for practical applications. 

Due to limited space,  a few aspects in face alignment have not been studied in this paper, e.g, what is the impact of training data to model performance? How scalable and extensible  of a method? How to enhance the robustness against initialisation variation? They are all interesting and we will investigate in our future work.

\bibliographystyle{ieee}
\bibliography{cvpr}
\end{document}